\documentclass[conference]{IEEEtran}
\IEEEoverridecommandlockouts
\usepackage{cite}
\usepackage{amsmath,amssymb,amsfonts}
\usepackage{textcomp}
\usepackage{algorithmic}
\usepackage{graphicx}
\usepackage{textcomp}
\usepackage{xcolor}
\def\BibTeX{{\rm B\kern-.05em{\sc i\kern-.025em b}\kern-.08em
    T\kern-.1667em\lower.7ex\hbox{E}\kern-.125emX}}
\begin{document}

\title{CP-EB: Talking Face Generation with Controllable Pose and Eye Blinking Embedding\\
}

\author{\IEEEauthorblockN{Jianzong Wang$^{1}$, Yimin Deng$^{1,2}$, Ziqi Liang$^{1,2}$, Xulong Zhang$^{1\ast}$\thanks{$^\ast$Corresponding author: Xulong Zhang (zhangxulong@ieee.org).}, Ning Cheng$^{1}$, Jing Xiao$^{1}$}
\IEEEauthorblockA{\textit{$^{1}$Ping An Technology (Shenzhen) Co., Ltd.}\\\textit{$^{2}$University of Science and Technology of China}}
}

\maketitle

\begin{abstract}
This paper proposes a talking face generation method named ``CP-EB'' that takes an audio signal as input and a person image as reference, to synthesize a photo-realistic people talking video with head poses controlled by a short video clip and proper eye blinking embedding. It's noted that not only the head pose but also eye blinking are both important aspects for deep fake detection. The implicit control of poses by video has already achieved by the state-of-art work. According to recent research, eye blinking has weak correlation with input audio which means eye blinks extraction from audio and generation are possible. Hence, we propose a GAN-based architecture to extract eye blink feature from input audio and reference video respectively and employ contrastive training between them, then embed it into the concatenated features of identity and poses to generate talking face images. Experimental results show that the proposed method can generate photo-realistic talking face with synchronous lips motions, natural head poses and blinking eyes.
\end{abstract}


\section{Introduction}
With the rapid development of deep learning techniques and smart devices, talking face generation has become a popular topic in computer vision and speech recognition. As metaverse has aroused a lot of attention, human avatar with talking face is a common way for people to express themselves in metaverse. In this paper, we aim at the task of talking face generation given a source image of target person and audio.

Talking face generation aims to create photo-realistic portraits~\cite{yu2020multimodal,wen2020photorealistic}. Given a person image as identity reference, one-shot talking face generation  synthesizes a person talking video by an input audio which seems like image animation. It can be considered as an image animation task \cite{siarohin2021motion}. To animate an input image, current methods can be divided into two categories, intermediate assist based and pure latent feature learning based~\cite{zhou2019talking,zhou2021pose}, depending on whether using intermediate representations like 2D landmarks \cite{chen2019hierarchical} and 3D morphable models \cite{zhang2021flow}. With the help of intermediate assist, explicit controlling for talking people can be realized. However, it means more categories of loss will be introduced. This paper intends to improve the performance of learning latent features from input sources with less intermediate loss.

Realistic talking face generation aims at driving a realistic people talking by input audio simultaneously, which is much more than the synchronization between lips and audio. It's easy for deep fake techniques to judge whether it's synthesized. According to recent research of deepfake~\cite{masood2022deepfakes,li2021exposing,yu2021survey}, most of exisiting detection approaches focus on whether there are a certain 
amount of proper physical features such as head poses, eye blinking, face warping, etc. To make the generated results vivid and convicible, we'd better provide natural head pose and blinking eyes in generation. Head reenactment \cite{burkov2020neural,doukas2021head2head++,tripathy2022single} focuses on head pose of talking people and aims to perform cross-person pose transfer. Zhou \emph{et al.}'s work \cite{zhou2021pose} use an input video as pose reference to control head motions of talking person. Nowadays, eye blinking has become an important part for current deep fake detection, related to speaker's habits and speech content. In the field of talking face, Liu \emph{et al.}~\cite{liu2022generating} generates lip motions and controllable blinking actions with fixed head poses. Most research on eye movement relies on driven-video. Recently, Zhang \emph{et al.}~\cite{zhang2021facial} proposes FACIAL GAN to extract eye blink features from audio and use it in rendering network as additional visual reference. And Chen's work~\cite{chen2022talking} also extract facial action units information from audio which covers eye area. Inspired by FACIAL GAN and PC-AVS, we decide to generate talking face with controllable head pose and blinking eyes in which contrastive learning is applied in pose and eye feature learning. 

\begin{figure*}[htb]
  \centering
  {\includegraphics[scale=0.4]{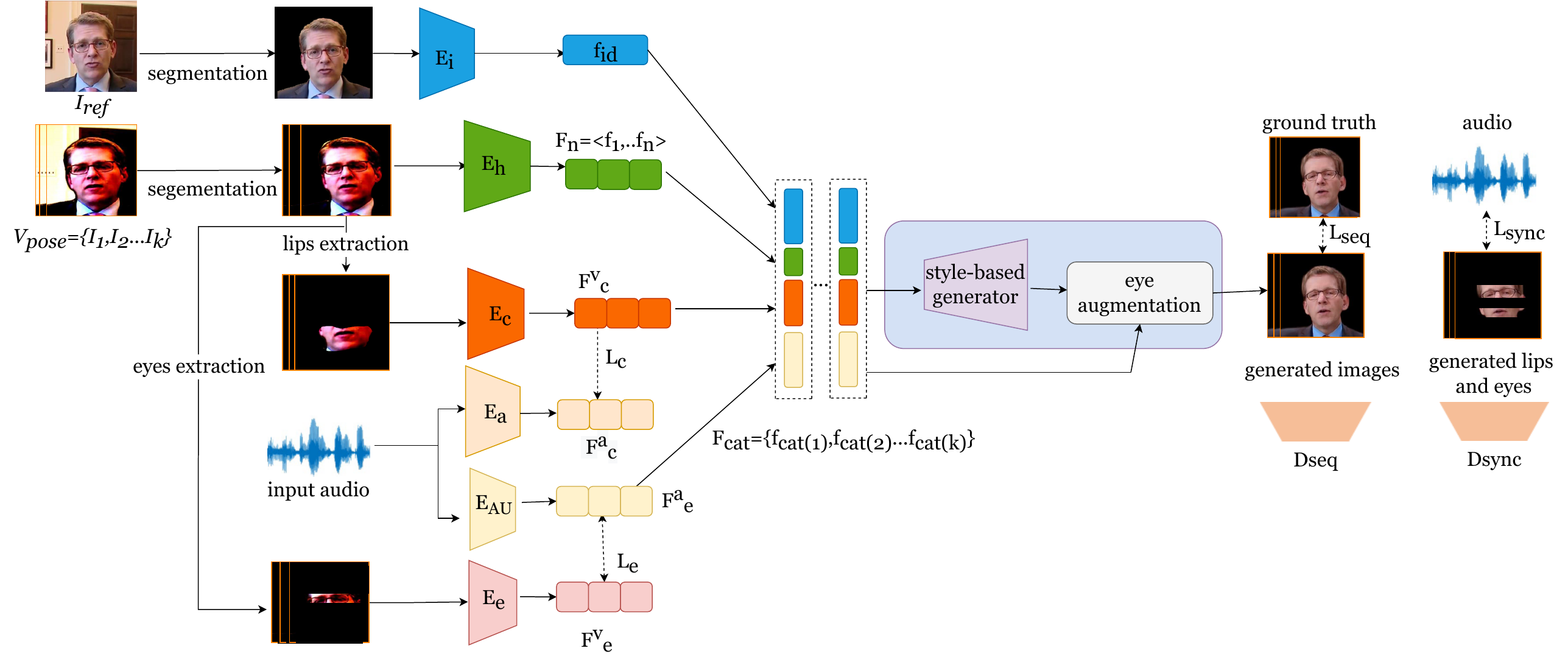}}
  \caption{\textbf{Pipeline of the proposed generation framework. }The encoders for image processing include identity encoder $E_i$, head pose encoder $E_h$, lips encoder for content $E_c$ and eye encoder $E_e$. The encoders for audio processing include audio-content $E_c$ and audio-face $E_{AU}$. Different features extracted from encoders are concatenated as $F_{cat}$ and fed into a style-based generator. Then the result is processed by an eye augmentation module. Two discriminators are designed with two kinds of loss to discriminate the result between ground truth images and audio in the aspects of naturalness and synchronization. }
  \label{fig:framework}
\end{figure*}

This paper proposes a talking face generation method named CP-EB which realizes pose controlling and eye blinking. Head pose is controlled by a pose encoder with reference input video. During the preprocess stage, CP-EB detects the eyes area of pose reference video and encode it into latent space. An  encoder-decoder network is designed for implicitly learning identity and pose information from input data. A combination of temporal encoder and location encoder is applied to extract eye blinking features from audio. Eyes blinking detection techniques are ultilized to detect blinking in video and extract corresponding features from video. Then use contrastive training for eye blinking latent vectors that should be in sync with input audio which then indicate the eyes transformation. As augmentation, eyes transformation is achieved by landmarks movement around eyes area which shows that eyes are blinking. Our contributions of this paper are summarized as follows:
\begin{itemize}
\item We improve audio-visual representation framework by embedding eye blinking. Contrastive learning is applied to eye blinking information from both video and audio to embed eye motions into talking face. 
\item To put more attention to target area, we apply segmentation at the preprocess stage before extracting head pose, mouth and eyes area from video frames. 
\item To improve the performance of classic implicit method in generation stage, we also introduce local landmarks prediction for augmentation. 
\end{itemize}

\section{Related Work}
\subsection{Talking Face Generation}
Given a person image as identity reference, one-shot talking face generation synthesizes a person talking video by an input audio~\cite{toshpulatov2023talking}. Such talking face generation requires producing realistic facial movements and synchronized speech in response to audio input. With the rapid evolution of deep learning, it becomes easily to handle with a huge amount of audio and visual data and producing satisfying results with techniques like Generative Adversarial Network~(GAN)~\cite{prajwal2020lip,fang2022facial} and diffusion model~\cite{zhua2023audio,zhang2022shallow}. Recent methods focus on the optimization on the important parts, such as identity preservation~\cite{bounareli2023stylemask}, face animation~\cite{ji2022eamm}, pose control~\cite{zhou2021pose} and audio-video synchronization~\cite{sun2022Pre-Avatar}.  

As controllable explicit features, 2D landmarks and 3DMM have been applied to face animation recent years. But such explicit features are inadequate for fine facial details and micro-expressions due to limited representation space. DAVS~\cite{zhou2019talking} first proposes a novel associative-and-adversarial training process for audio and video in latent space with implicit modeling. Proper head pose can pass the deep fake detection. Neural head reenactment~\cite{burkov2020neural} aims to control the movement of human head with latent pose representation. It encourages the encoder to implicit learn the head movement information with proper fore-ground segmentation. Moreover, PC-AVS~\cite{zhou2021pose} proposes another implicit modeling for talking face generation using a pose-reference video to control the head pose.   

\subsection{Granular Control for Expressive Generation}
Furthermore, recent methods based on latent feature learning focus on various emotional expression. GC-AVT~\cite{liang2022expressive} proposes a novel prior-based pre-processing for different facial parts to drive portrait head with a higher level of granularity. For facial details, actually the facial dynamics are dominated by the shape of the mouth and eyes. We need to explore a more efficient and precise control for generation. Previous work has already made a huge progress in lip-sync area~\cite{prajwal2020lip} while the movements of eyes still need further research.

Eye blinking is an important signal in communication. Hao's work~\cite{hao2021controlling} learns the eye blinking feature from video with eye state labels. APB2Face~\cite{zhang2020apb2face} detects and reconstructs the eye states by geometry predictor. Besides, the synchronization about eyes deserves exploring. FACIAL-GAN~\cite{zhang2021facial} explores the weak correlation between eyes and audio which is potential to improve the dynamic modeling of eye area in talking face generation. Inspired by this, we can extend the basic talking face generation to expressive eye blinking learning features from video and audio.

\section{Methodology}

\subsection{Identity Encoder}
The identity encoder $E_i$ takes a person image as input and gives out the identity feature. When in the training process, CP-EB makes the identity features of different images of the same input person as similar as possible. In reference, CP-EB randomly samples one frame of given video as identity reference, and encodes the identity into latent space.

\subsection{Head pose Encoder}
The head pose encoder mostly learns identity-dependent information from consecutive video frames including head pose and lips. In head pose learning, CP-EB uses the early work of \cite{zhou2019talking} to establish pose encoder $E_h$ including HourGlass-like network. To strengthen the ability of representation, contrastive learning is used in cross-modal training process. For lip synthesis, CP-EB learns a visual-content encoder $E_c$ and an audio-content encoder $E_a$. We take the corresponding audio as input and then convert it to mel-spectrograms which are then sent to a ResNet-based encoder and transfer into content space. We use contrastive training between audio-content feature and visual-content feature with the contrastive loss. Cosine distance $Dis_{c}$ is appled in computing similarity between them.
\begin{equation}
    \begin{split}
    L_{aud}^{vid} =
    -log[\frac{exp(Dis_{c}(F^v_c,F^{a^{P}}_c))}{\sum^{S}_{i=1}exp(Dis_{c}(F^v_c,F^{a}_{c_{i}}))}]
    \end{split}
\end{equation}

\par where $S$ means the total number of positive and negative samples, the superscript $(\cdot)^P$ indicates the positive sample related to the task.
\par The audio to visual loss $L^{a}_{v}$ can be formulated in a symmetric way. The total loss can be organized as:
\begin{equation}
    \begin{split}
    L_{c}=L_{aud}^{vid}+L^{aud}_{vid}
    \end{split}
\end{equation}

\subsection{Eyes blinking Encoder}
\par Proper eyes blinking helps to eliminate uncanny valley effect in deep fake detection. CP-EB learns the movement of specific facial action units (FACS) \cite{ekman1978facial} to represent the stage of eyes. FACS scores individual and interrelated motor units to represent different phases called AU values. Compared to 68 facial landmarks, using action of facial units as representation of eyes stage not only reduces complexity and intermediate loss but also is consistent with physiological principle. In the following steps, this value will be considered as indicator for eyes transformation. The eye blinking encoders are trained in a supervised manner. To make paired data, we utilize OpenFace \cite{baltrusaitis2018openface} to provide groundtruth which captures facial actions with landmarks from videos. In OpenFace, the intensity of eye blinking action is measured on a scale of 0 to 5 and the values are saved in a csv file. 
\begin{figure}[htb]
  \centering
  {\includegraphics[scale=0.13]{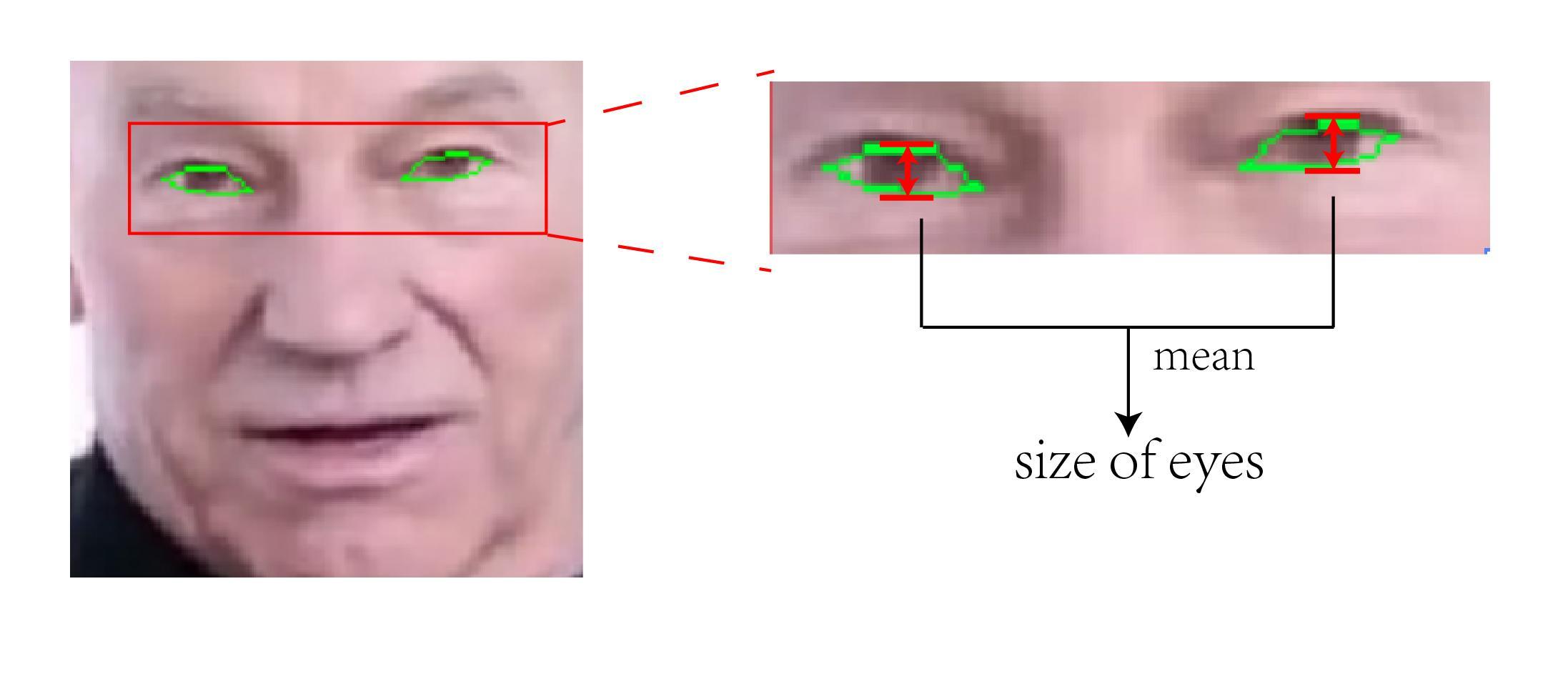}}
  \caption{\textbf{Visualization of eyes detection and the definition of eyes size }  }
  \label{fig:eyesize}
\end{figure}
\par It's proved that there is weak correlated information between blinkings and DeepSpeech features \cite{hannun2014deep} of audio \cite{zhang2021facial}. To extract corresponding AU value in each audio feature frame, we combine temporal features of the whole audio and local features of each frames and map into a AU-related vector. In visual work, we extract relative features from images by masking eye and train a classifier to predict AU45 value on a scale of 0 to 5 in consecutive frames. Contrastive learning is also adopted in learning eye blinking representation. We minimize L2-form contrasive learning loss to maximize the similarity between audio-blink and video-blink representation. Finally, we concatenate the identity, head pose, lips and eye blinking features as the final implicit representation in latent space and send them into generator.

\subsection{Generation with Discriminators}
In this subsection, we will describe the structure of generator and discriminator and blinking eyes embedding. Adaptive instance normalization (AdaIN) is proved to perform well in image generation. So we apply a style-based generator which takes the concatenated features as input in each layer. We design the use of MLP and AdaIN block to process the original input before injecting it into each layer. We found that implicit method is weak in directly generating blinking eye. So we put AU-related features as indicator for the following augmentation of eye generation.

\begin{figure}[ht]
  \centering
  {\includegraphics[scale=0.6]{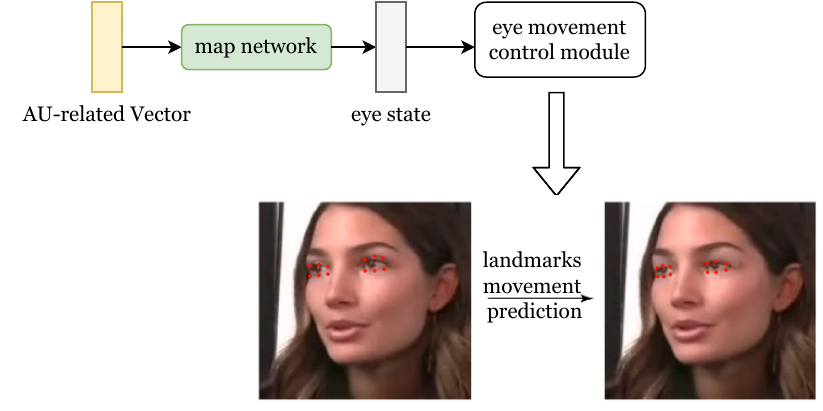}}
  \caption{\textbf{Realization of eye augmentation }  }
  \label{fig:blink}
\end{figure}

Specifically, to realize eyes transformation, we extract the AU-related vector from input concatenated features and map into the stage of eyes which indicates landmarks movement. Eyes stage indicates the size of eyes in current frame.  As shown in Fig.~\ref{fig:eyesize}, it's common that both eyes blinks in the same time, so we take the mean size of two eyes as definition of current size of both eyes. In other words, we try to move eyes landmarks with similar ratio. There are 68 facial landmarks in total while we only move the landmarks around eyes area. As shown in Fig.~\ref{fig:blink}, each side involves 6 landmarks. We design a mapping network with convolution layers and a fully connection layer. Take AU-related vector as input, this mapping network learns current size of eyes which is then fed into control module. This module predicts related landmarks movement for eyes transformation. 

In the stage of discrimination, we design a reconstruction discriminator and synchronization discriminator. Loss function of GAN can be formulated as follows: 
\begin{equation}
  \begin{split}
    L_{GAN} = &arg\min\limits_{G}\max\limits_{D}\sum^{N_D}_{n=1}(\mathbb{E}_{I_k}[\log D_n(I_k)]\\
    &+\mathbb{E}_{I_k}[\log(1-D_n(G(f_{cat}{(k)}))])
\end{split}
\end{equation}
\par Pixel-wise $L_1$ distances between each groundtruth image
$I_k$ and synthesized image $G(f_{cat(k)})$ are leveraged in $D_{seq}$ to learn reconstruction objective. We also use pretrained VGG network to compute perceptual loss \cite{johnson2016perceptual}.
\begin{equation}
  \begin{split}
    L_1 = ||I_k-G(f_{cat}{(k)})||_1
\end{split}
\end{equation}
\begin{equation}
  \begin{split}
    L_{VGG} = &\sum^{N_P}_{n-1}||VGG_n(I_k)-VGG_n(G(f_{cat}{(k)}))||_1
\end{split}
\end{equation}
\par We modify a lip-sync discriminator $D_{sync}$ from \cite{wang2022one} and extend to eye-sync measurement to compute synchronization loss between image and data. We adopt SyncNet in Wav2Lip \cite{prajwal2020lip} to extract visual embedding including lips embedding $e_{lips}$ and eyes embedding $e_{eyes}$ from video frames and audio embedding $e_a$ from audio. The synchronization probability is showed by calculating the cosine distance between them. The negative 
logarithm form of this probability is used to formulate the loss of synchronization $L_{sync}$.
\begin{equation}
  \begin{split}
    P_{sync} = \frac{(e_{lips} \oplus e_{eyes}) \cdot e_a}{\mathop{max(||(e_{lips} \oplus e_{eyes})||_2 \cdot ||e_a||_2,\epsilon)}}
\end{split}
\end{equation}
\begin{equation}
\begin{split}
  L_{sync} = -\log{P_{sync}}
\end{split}
\end{equation}
\par The overall loss functions during training can be summarized as:
\begin{equation}
  \begin{split}
    L_{all} = L_{GAN} + \lambda_1L_{1} + \lambda_2L_{VGG} + \lambda_3L_{sync}
\end{split}
\end{equation}
\par where $\lambda_1$, $\lambda_2$ and $\lambda_3$ are hyperparameters.

\begin{figure*}[htb]
  \centering
  {\includegraphics[scale=0.17]{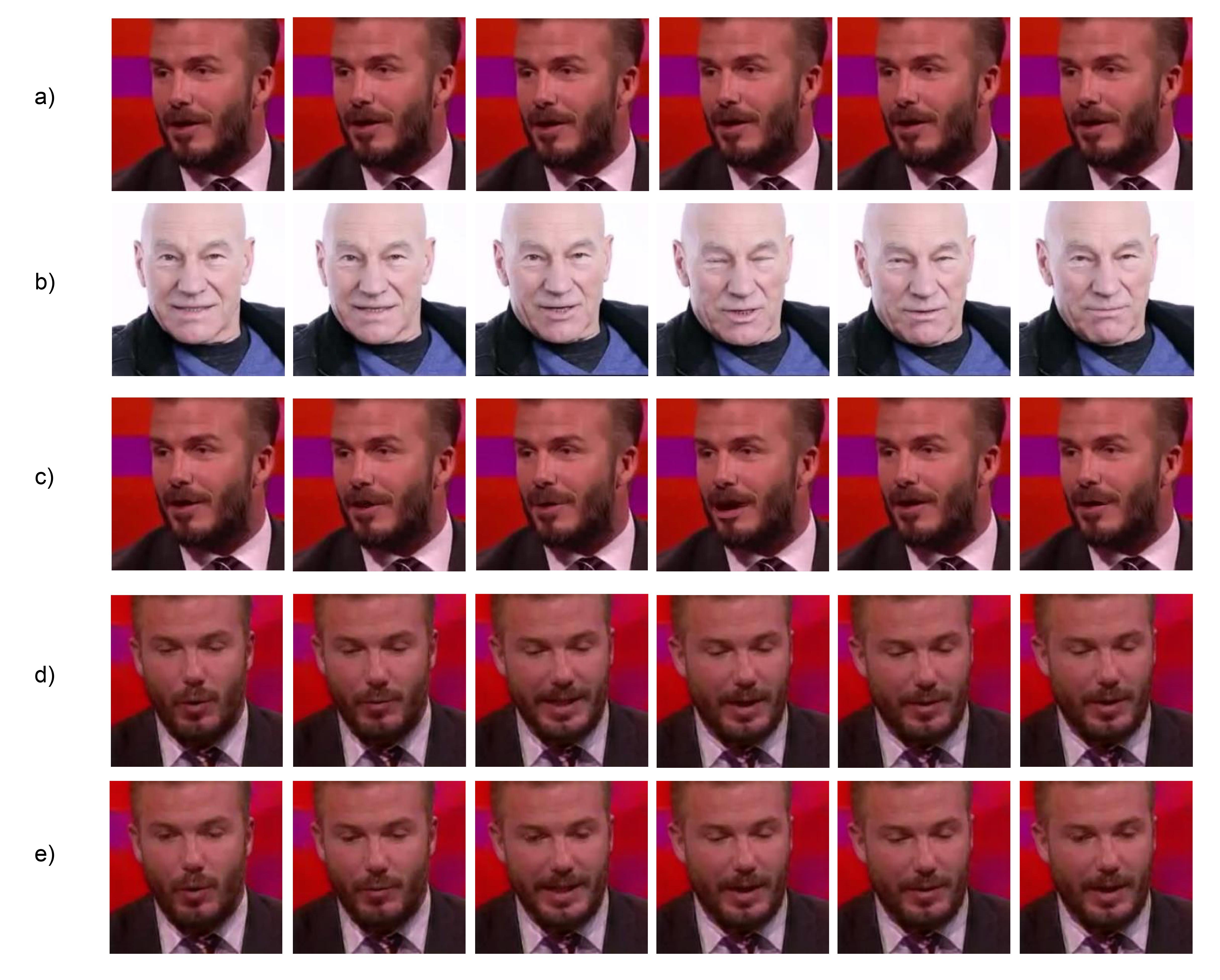}}
  \caption{\textbf{Comparison with baseline methods for talking face generation. }Group a) is listed as the identity reference. Group b) is used to provide pose reference. The following results group c) and group d) are from DAVS and PC-AVS respectively. The bottom group e) is from proposed method CP-EB which achieves head pose changing and eyes blinking.  }
  \label{fig:exp}
\end{figure*}


\section{Experiments and Results}
\subsection{Dataset}
The total model is learned on video data collected from VoxCeleb2 \cite{chung2018voxceleb2} video dataset collected from Youtube and cropped into a fixed size. The whole dataset contains more than a million videos from over 6000 speakers in different languages. For our model training and testing, we make use of a subset of Voxcecleb2. Each raw sequence includes one person who talks for few seconds on various occasions. There are 430 persons in total in our dataset. The dataset is divided into a training set and a testing set at a ratio of 8 to 2, choosing 344 people for training and 86 people for testing.

\subsection{Experimental setup}
Proper data augmentation helps to remove redundant information and improve model's performance \cite{zhou2021pose}. Color transfer is applied in extracting identity-dependent features in our work. Pose reference video frames are pre-processed by randomly altering RGB channels. We also use dlib library for face detection which mainly detects our target area (mouth and eyes) to get corresponding masks which are then fed into encoders. 
\par We utilize an architecture of 6 convolution layers to extract eye blinking feature from audio features to extract it from eye video frames. In visual network, identity encoders and lips-related encoders are similar to PC-AVS. In audio-blink network, we design a temporal encoder to capture features of the whole audio and use a local encoder to capture features of each audio frame. Assumed that there are K frames to process, we need a temporal encoder and K local encoders for feature extraction. So the temporal features can be divided into K parts and then concatenated with local features. Then the features pass through a fully connected layer and output a  71-dimension vector which predicts the AU45 value in 6th dimension. In video-blink network, we adopt eye stage classifier \cite{liu2022generating} in our video-blink network and pretrain it with our dataset. In details, we use dlib to detect the landmarks movement of eyes in frames and get current size of eyes from eye stage classifier. And then map it into AU45 value. We use 4 Tesla V100 to accelerate the whole work for a week. To generate a complete result, in the pre-process stage we save the background information and add it in the final step.  

\subsection{Evaluation}
As for evaluation, we choose SSIM (Structural Similarity) to evaluate the reconstruction quality which is higher for better. We also compute Landmarks Distance around mouths  ($\boldsymbol{\rm{LMD}_{mouth}}$) \cite{chen2019hierarchical} and extend it to eyes($\boldsymbol{\rm{LMD}_{eye}}$) to evaluate the accuracy of mouth shapes and eyes blinking which are lower for better. During 68 facial landmarks, there are 6 landmarks around each eye and 8 landmarks around mouth.
Former works in the field of talking face are selected as baseline including DAVS~\cite{zhou2019talking}, PC-AVS \cite{zhou2021pose}. We also conduct an ablation study with/without eye blinking embedding.

\subsection{Results}
\textbf{Qualitative Results} We compare generated results from our methods against previous talking face methods DAVS and PC-AVS. As we can see in Fig.~\ref{fig:exp}, row a) provides an identity reference whose head turns left. While row b) provides a video for pose reference in which the man is looking straight ahead. Row c) shows the results from DAVS which only change the mouth area and 
keep the rest parts same. Row d) shows the results from PC-AVS which change the head pose but the eyes keep close all the time. The results of proposed methods are showed in row d) which also change head pose according to pose reference video. What's more, it's seen that some transformations appear around the eyes area and the size of eyes is not fixed.

\textbf{Quantitative Results} In quantitative tests, we select one more baseline model FAICIAL~\cite{zhang2021facial} to make comparison. Since it utilizes 3D model guidance, we show the gap between our implicit modeling capabilities and hybrid modeling through comparison. We evaluate our methods with baseline works using the chosen metrics $SSIM$ and $LMD$. As mentioned in former subsection, $SSIM$ is higher for better while $LMD$ is lower for better. As shown in Table \ref{table_1}, compared to the baseline models, our method can achieve a lower value of ${LMD}_{eye}$ and higher score of $SSIM$ and ${LMD}_{mouth}$. However, FACIAL performs better in eye blinking modeling due to the guidance of the explicit features. The performance of CP-EB based on implicit modeling has strived to approximate this effect. Both qualitative results and quantitative results indicate that our model improves the comprehensive modeling ability in talking face generation. 

\begin{table}[ht]
  \renewcommand{\arraystretch}{1.2}
  \caption{Quantitative results on subset of VoxCeleb2}
  \label{table_1}
  \centering
  \begin{tabular}{c c c c}
      \hline
       & $\boldsymbol{\rm{SSIM}}$${\uparrow}$ & $\boldsymbol{\rm{LMD}_{mouth}{\downarrow}}$ & $\boldsymbol{\rm{LMD}_{eye}{\downarrow}}$  \\
      \hline
      PC-AVS~\cite{zhou2021pose} &0.717  &1.562  &5.124   \\
      DAVS~\cite{zhou2019talking} &0.622  &2.557  &7.920   \\
      FACIAL~\cite{zhang2021facial} &0.703  &1.550  &3.508  \\
      CP-EB &0.721  &1.547  &3.826   \\
      \hline
  \end{tabular}
\end{table}

\textbf{Ablation Study} We conduct ablation experiment by removing pose control information and eyes control information. As for head pose, we smooth the pose feature before feed it into generator. As for eye blinking generation, we need to analyze the AU-related vector and augmentation part. One is to smooth AU-related vector before input to generator while fixing the latter landmarks movement of which the input cannot provide useful indication. Another is removing augmentation for landmarks prediction. In Table \ref{table_2}, we show the results of these ablation studies. As mentioned in former subsection, $SSIM$ is higher for better while $LMD$ is lower for better.

As we can see in Table \ref{table_2}, it's proved that pose module and eyes module are efficient for quality improvement. As for eyes controlling, AU-related information is of a certain importance for blinking eye generation. Besides, eyes transformation with landmarks is useful as augmentation for implicit generation in our method.
\begin{table}[ht]
  \renewcommand{\arraystretch}{1.2}
  
  \caption{Quantitative results for ablation studies}
  \label{table_2}
  \centering
  \begin{tabular}{c c c c}
      \hline
       & $\boldsymbol{\rm{SSIM}}$${\uparrow}$ & $\boldsymbol{\rm{LMD}_{mouth}{\downarrow}}$ & $\boldsymbol{\rm{LMD}_{eye}{\downarrow}}$  \\
      \hline
      CP-EB  &0.689  &1.486  &3.048   \\
      w/o pose &0.437  &1.535  &3.283   \\
      w/o AU &0.628  &1.479  &5.500   \\
      w/o eye aug &0.596  &1.481  &4.921 \\
      \hline
  \end{tabular}
\end{table}
\par\textbf{Limitation} Though we realize eye blinking embedding, the change of eyes is not as much obvious as possible. And the eyes transformation is much related to the information delivered by AU value of eyes units which means little intermediate loss is introduced.


\section{Conclusion}
We propose a method to control the head pose and embed blinking eyes in talking face generation in the same time. Implict learning from talking videos is an effective way to reduce intermediate loss during generation. We extend the implicit method in eye area generation by making full use of the weak information in audio and eye blink detection technique in video. The experiments shows that we can realize a more vivid result than baseline work. It's worth noting that expressive results deserve further research in implicit talking face generation.

\section{Acknowledgment}
Supported by the Key Research and Development Pro- gram of Guangdong Province (grant No. 2021B0101400003) and corresponding author is Xulong Zhang (zhangxulong@ieee.org).




\bibliographystyle{IEEEtran.bst}
\bibliography{mybib.bib}

\begin{thebibliography}{10}
\providecommand{\url}[1]{#1}
\csname url@samestyle\endcsname
\providecommand{\newblock}{\relax}
\providecommand{\bibinfo}[2]{#2}
\providecommand{\BIBentrySTDinterwordspacing}{\spaceskip=0pt\relax}
\providecommand{\BIBentryALTinterwordstretchfactor}{4}
\providecommand{\BIBentryALTinterwordspacing}{\spaceskip=\fontdimen2\font plus
\BIBentryALTinterwordstretchfactor\fontdimen3\font minus \fontdimen4\font\relax}
\providecommand{\BIBforeignlanguage}[2]{{%
\expandafter\ifx\csname l@#1\endcsname\relax
\typeout{** WARNING: IEEEtran.bst: No hyphenation pattern has been}%
\typeout{** loaded for the language `#1'. Using the pattern for}%
\typeout{** the default language instead.}%
\else
\language=\csname l@#1\endcsname
\fi
#2}}
\providecommand{\BIBdecl}{\relax}
\BIBdecl

\bibitem{yu2020multimodal}
L.~Yu, J.~Yu, M.~Li, and Q.~Ling, ``Multimodal inputs driven talking face generation with spatial--temporal dependency,'' \emph{IEEE Transactions on Circuits and Systems for Video Technology}, vol.~31, no.~1, pp. 203--216, 2020.

\bibitem{wen2020photorealistic}
X.~Wen, M.~Wang, C.~Richardt, Z.-Y. Chen, and S.-M. Hu, ``Photorealistic audio-driven video portraits,'' \emph{IEEE Transactions on Visualization and Computer Graphics}, vol.~26, no.~12, pp. 3457--3466, 2020.

\bibitem{siarohin2021motion}
A.~Siarohin, O.~J. Woodford, J.~Ren, M.~Chai, and S.~Tulyakov, ``Motion representations for articulated animation,'' in \emph{Proceedings of the IEEE/CVF Conference on Computer Vision and Pattern Recognition (CVPR)}, 2021, pp. 13\,653--13\,662.

\bibitem{zhou2019talking}
H.~Zhou, Y.~Liu, Z.~Liu, P.~Luo, and X.~Wang, ``Talking face generation by adversarially disentangled audio-visual representation,'' in \emph{Proceedings of the AAAI conference on artificial intelligence}, 2019, pp. 9299--9306.

\bibitem{zhou2021pose}
H.~Zhou, Y.~Sun, W.~Wu, C.~C. Loy, X.~Wang, and Z.~Liu, ``Pose-controllable talking face generation by implicitly modularized audio-visual representation,'' in \emph{Proceedings of the IEEE/CVF conference on Computer Vision and Pattern Recognition (CVPR)}, 2021, pp. 4176--4186.

\bibitem{chen2019hierarchical}
L.~Chen, R.~K. Maddox, Z.~Duan, and C.~Xu, ``Hierarchical cross-modal talking face generation with dynamic pixel-wise loss,'' in \emph{Proceedings of the IEEE/CVF conference on Computer Vision and Pattern Recognition (CVPR)}, 2019, pp. 7832--7841.

\bibitem{zhang2021flow}
Z.~Zhang, L.~Li, Y.~Ding, and C.~Fan, ``Flow-guided one-shot talking face generation with a high-resolution audio-visual dataset,'' in \emph{Proceedings of the IEEE/CVF Conference on Computer Vision and Pattern Recognition (CVPR)}, 2021, pp. 3661--3670.

\bibitem{masood2022deepfakes}
M.~Masood, M.~Nawaz, K.~M. Malik, A.~Javed, A.~Irtaza, and H.~Malik, ``Deepfakes generation and detection: State-of-the-art, open challenges, countermeasures, and way forward,'' \emph{Applied Intelligence}, pp. 1--53, 2022.

\bibitem{li2021exposing}
M.~Li, B.~Liu, Y.~Hu, and Y.~Wang, ``Exposing deepfake videos by tracking eye movements,'' in \emph{2020 25th International Conference on Pattern Recognition (ICPR)}.\hskip 1em plus 0.5em minus 0.4em\relax IEEE, 2021, pp. 5184--5189.

\bibitem{yu2021survey}
P.~Yu, Z.~Xia, J.~Fei, and Y.~Lu, ``A survey on deepfake video detection,'' \emph{Iet Biometrics}, vol.~10, no.~6, pp. 607--624, 2021.

\bibitem{burkov2020neural}
E.~Burkov, I.~Pasechnik, A.~Grigorev, and V.~Lempitsky, ``Neural head reenactment with latent pose descriptors,'' in \emph{Proceedings of the IEEE/CVF conference on Computer Vision and Pattern Recognition (CVPR)}, 2020, pp. 13\,786--13\,795.

\bibitem{doukas2021head2head++}
M.~C. Doukas, M.~R. Koujan, V.~Sharmanska, A.~Roussos, and S.~Zafeiriou, ``Head2head++: Deep facial attributes re-targeting,'' \emph{IEEE Transactions on Biometrics, Behavior, and Identity Science}, vol.~3, no.~1, pp. 31--43, 2021.

\bibitem{tripathy2022single}
S.~Tripathy, J.~Kannala, and E.~Rahtu, ``Single source one shot reenactment using weighted motion from paired feature points,'' in \emph{Proceedings of the IEEE/CVF Winter Conference on Applications of Computer Vision}, 2022, pp. 2715--2724.

\bibitem{liu2022generating}
S.~Liu and J.~Hao, ``Generating talking face with controllable eye movements by disentangled blinking feature,'' \emph{IEEE Transactions on Visualization \& Computer Graphics}, no.~01, pp. 1--12, 2022.

\bibitem{zhang2021facial}
C.~Zhang, Y.~Zhao, Y.~Huang, M.~Zeng, S.~Ni, M.~Budagavi, and X.~Guo, ``Facial: Synthesizing dynamic talking face with implicit attribute learning,'' in \emph{Proceedings of the IEEE/CVF International Conference on Computer Vision}, 2021, pp. 3867--3876.

\bibitem{chen2022talking}
S.~Chen, Z.~Liu, J.~Liu, and L.~Wang, ``Talking head generation driven by speech-related facial action units and audio-based on multimodal representation fusion,'' \emph{CoRR}, vol. abs/2204.12756, 2022.

\bibitem{toshpulatov2023talking}
M.~Toshpulatov, W.~Lee, and S.~Lee, ``Talking human face generation: A survey,'' \emph{Expert Systems with Applications}, p. 119678, 2023.

\bibitem{prajwal2020lip}
K.~Prajwal, R.~Mukhopadhyay, V.~P. Namboodiri, and C.~Jawahar, ``A lip sync expert is all you need for speech to lip generation in the wild,'' in \emph{Proceedings of the 28th ACM International Conference on Multimedia}, 2020, pp. 484--492.

\bibitem{fang2022facial}
Z.~Fang, Z.~Liu, T.~Liu, C.-C. Hung, J.~Xiao, and G.~Feng, ``Facial expression gan for voice-driven face generation,'' \emph{The Visual Computer}, pp. 1--14, 2022.

\bibitem{zhua2023audio}
Y.~Zhua, C.~Zhanga, Q.~Liub, and X.~Zhoub, ``Audio-driven talking head video generation with diffusion model,'' in \emph{ICASSP 2023-2023 IEEE International Conference on Acoustics, Speech and Signal Processing (ICASSP)}, 2023, pp. 1--5.

\bibitem{zhang2022shallow}
X.~Zhang, J.~Wang, N.~Cheng, E.~Xiao, and J.~Xiao, ``Shallow diffusion motion model for talking face generation from speech,'' in \emph{Asia-Pacific Web (APWeb) and Web-Age Information Management (WAIM) Joint International Conference on Web and Big Data}, 2022, pp. 144--157.

\bibitem{bounareli2023stylemask}
S.~Bounareli, C.~Tzelepis, V.~Argyriou, I.~Patras, and G.~Tzimiropoulos, ``Stylemask: Disentangling the style space of stylegan2 for neural face reenactment,'' in \emph{2023 IEEE 17th International Conference on Automatic Face and Gesture Recognition (FG)}, 2023, pp. 1--8.

\bibitem{ji2022eamm}
X.~Ji, H.~Zhou, K.~Wang, Q.~Wu, W.~Wu, F.~Xu, and X.~Cao, ``Eamm: One-shot emotional talking face via audio-based emotion-aware motion model,'' in \emph{{SIGGRAPH} '22: Special Interest Group on Computer Graphics and Interactive Techniques Conference, Vancouver, BC, Canada, August 7 - 11, 2022}, 2022, pp. 1--10.

\bibitem{sun2022Pre-Avatar}
A.~Sun, X.~Zhang, T.~Ling, J.~Wang, N.~Cheng, and J.~Xiao, ``Pre-avatar: An automatic presentation generation framework leveraging talking avatar,'' in \emph{2022 IEEE 34th International Conference on Tools with Artificial Intelligence (ICTAI)}, 2022, pp. 1002--1006.

\bibitem{liang2022expressive}
B.~Liang, Y.~Pan, Z.~Guo, H.~Zhou, Z.~Hong, X.~Han, J.~Han, J.~Liu, E.~Ding, and J.~Wang, ``Expressive talking head generation with granular audio-visual control,'' in \emph{Proceedings of the IEEE/CVF Conference on Computer Vision and Pattern Recognition (CVPR)}, 2022, pp. 3387--3396.

\bibitem{hao2021controlling}
J.~Hao, S.~Liu, and Q.~Xu, ``Controlling eye blink for talking face generation via eye conversion,'' in \emph{{SA} '21: {SIGGRAPH} Asia 2021 Technical Communications, Tokyo, Japan, December 14 - 17, 2021}, 2021, pp. 1--4.

\bibitem{zhang2020apb2face}
J.~Zhang, L.~Liu, Z.~Xue, and Y.~Liu, ``Apb2face: Audio-guided face reenactment with auxiliary pose and blink signals,'' in \emph{ICASSP 2020-2020 IEEE International Conference on Acoustics, Speech and Signal Processing (ICASSP)}.\hskip 1em plus 0.5em minus 0.4em\relax IEEE, 2020, pp. 4402--4406.

\bibitem{ekman1978facial}
P.~Ekman and W.~V. Friesen, ``Facial action coding system,'' \emph{Environmental Psychology \& Nonverbal Behavior}, 1978.

\bibitem{baltrusaitis2018openface}
T.~Baltrusaitis, A.~Zadeh, Y.~C. Lim, and L.-P. Morency, ``Openface 2.0: Facial behavior analysis toolkit,'' in \emph{2018 13th IEEE international conference on automatic face \& gesture recognition (FG 2018)}, 2018, pp. 59--66.

\bibitem{hannun2014deep}
A.~Y. Hannun, C.~Case, J.~Casper, B.~Catanzaro, G.~Diamos, E.~Elsen, R.~Prenger, S.~Satheesh, S.~Sengupta, A.~Coates, and A.~Y. Ng, ``Deep speech: Scaling up end-to-end speech recognition,'' \emph{CoRR}, vol. abs/1412.5567, 2014.

\bibitem{johnson2016perceptual}
J.~Johnson, A.~Alahi, and L.~Fei-Fei, ``Perceptual losses for real-time style transfer and super-resolution,'' in \emph{Computer Vision - {ECCV} 2016 - 14th European Conference, Amsterdam, The Netherlands, October 11-14, 2016, Proceedings, Part {II}}, 2016, pp. 694--711.

\bibitem{wang2022one}
S.~Wang, L.~Li, Y.~Ding, and X.~Yu, ``One-shot talking face generation from single-speaker audio-visual correlation learning,'' in \emph{Proceedings of the AAAI Conference on Artificial Intelligence}, 2022, pp. 2531--2539.

\bibitem{chung2018voxceleb2}
J.~S. Chung, A.~Nagrani, and A.~Zisserman, ``Voxceleb2: Deep speaker recognition,'' in \emph{Interspeech 2018, 19th Annual Conference of the International Speech Communication Association, Hyderabad, India, 2-6 September 2018}, 2018, pp. 1086--1090.

\end{thebibliography}

\end{document}